%% file: pro.tex
\documentclass[letterpaper,twocolumn,10pt]{article}
\usepackage{usenix-2020-09}

\input{preamble}

\begin{document}
\date{}

\title{Predictive Response Optimization: \\ Using Reinforcement Learning to Fight Online Social Network Abuse}

\author{
    {\rm Garrett Wilson}\\
    Meta Platforms, Inc.
    \and
    {\rm Geoffrey Goh}\\
    Meta Platforms, Inc.
    \and
    {\rm Yan Jiang}\\
    Meta Platforms, Inc.
    \and
    {\rm Ajay Gupta}\\
    Meta Platforms, Inc.
    \and
    {\rm Jiaxuan Wang}\\
    Meta Platforms, Inc.
    \and
    {\rm David Freeman}\\
    Meta Platforms, Inc.
    \and
    {\rm Francesco Dinuzzo}\\
    Meta Platforms, Inc.
}
\maketitle

\input{abstract}
\input{section1_intro}
\input{section2_background}
\input{section3_problem}
\input{section4_rl}
\input{section5_experiments}
\input{section7_ethics}
\input{section6_conclusion}

\bibliographystyle{plain}
\bibliography{bibliography}

\end{document}

%% file: preamble.tex
\newif\ifdraft
\draftfalse

\newif\ifanon
\anonfalse

\newif\ifsqueeze
\squeezetrue

\ifdraft
	\newcommand{\authorcomment}[3]{\textcolor{#2}{\noindent[#1: #3]}}
	\newcommand{\todo}[1]{\textcolor{red}{TODO: #1}}
	\newcommand{\owner}[1]{\textcolor{red}{\noindent[Owner: \textbf{#1}]}}
\else
	\newcommand{\authorcomment}[3]{}
	\newcommand{\todo}[1]{}
	\newcommand{\owner}[1]{}
\fi

\microtypecontext{spacing=nonfrench}

\newcommand{\squeezelist}{\ifsqueeze \setlength\itemsep{-2pt} \fi}

\newcommand{\ignore}[1]{}

\newcommand{\myparagraph}[1]{\ifsqueeze \smallskip \else \medskip \fi \noindent {\bf #1}}

\usepackage{algorithm}
\usepackage{algpseudocode}
\usepackage{amsfonts}
\usepackage{amsmath}
\usepackage{array}
\usepackage{booktabs}
\usepackage{caption}
\usepackage{graphicx}
\usepackage{subcaption}
\usepackage{xcolor}
\usepackage{xspace}
\usepackage{enumitem}

\newcommand{\fb}{\ifanon Social Network B\@\xspace\else Facebook\@\xspace\fi}
\newcommand{\ig}{\ifanon Social Network A\@\xspace\else Instagram\@\xspace\fi}
\newcommand{\fbig}{\ifanon two major social networks\@\xspace\else Instagram and Facebook\@\xspace\fi}
\newcommand{\osns}{OSNs\@\xspace}
\newcommand{\osn}{OSN\@\xspace}
\newcommand{\surface}{Mobile Web\@\xspace}

\newcommand{\smscost}{{\em dollars spent on SMS messaging}\@\xspace}
\newcommand{\scraping}{{\em weighted scraping requests}\@\xspace}
\newcommand{\automation}{{\em automated requests}\@\xspace}
\newcommand{\timespent}{{\em time spent}\@\xspace}
\newcommand{\daysactive}{{\em days active}\@\xspace}
\newcommand{\feedback}{{\em feedback events}\@\xspace}

\newcommand{\fbreduction}{4.5\%\@\xspace}
\newcommand{\igreduction}{59\%\@\xspace}
\newcommand{\fbdays}{14\@\xspace}
\newcommand{\igdays}{14\@\xspace}
\newcommand{\smsreduction}{80\%\@\xspace}
\newcommand{\mwebdaulossreduction}{68\%\@\xspace}
\newcommand{\mwebscrapingreduction}{12\%\@\xspace}

\newcommand{\newresponsereduction}{3.0\%\@\xspace}

\DeclareMathOperator*{\argmax}{\arg\!\max\;}

\algnewcommand\algorithmicforeach{\textbf{for each}}
\algdef{S}[FOR]{ForEach}[1]{\algorithmicforeach\ #1\ \algorithmicdo}

\newcommand{\AAA}{\mathcal{A}^{\abs{\mathcal{E}}}}

\newcommand{\abs}[1]{\left|#1\right|}

%% file: abstract.tex
\begin{abstract}

Detecting phishing, spam, fake accounts, data scraping, and other malicious
activity in online social networks (\osns) is a problem that has been studied
for well over a decade, with a number of important results. Nearly all
existing works on abuse detection have as their goal producing the best
possible binary classifier; i.e., one that labels unseen examples as
``benign'' or ``malicious'' with high precision and recall.
However, no prior published work considers what comes
next: what does the service actually {\em do} after it detects abuse?

In this paper, we argue that {\em detection} as described in previous work
{\em is not the goal} of those who are fighting \osn abuse. Rather, we
believe the goal to be {\em selecting actions} (e.g., ban the user, block the
request, show a CAPTCHA, or ``collect more evidence'') that {\em optimize a
tradeoff} between harm caused by abuse and impact on benign users.  With this
framing, we see that enlarging the set of possible actions allows us to move
the Pareto frontier in a way that is unattainable by simply tuning the
threshold of a binary classifier.

To demonstrate the potential of our approach, we present {\em Predictive
Response Optimization (PRO),} a system based on reinforcement learning that
utilizes available contextual information to predict future abuse and
user-experience metrics conditioned on each possible action, and select
actions that optimize a multi-dimensional tradeoff between abuse/harm and
impact on user experience.

We deployed versions of PRO targeted at stopping automated activity on \fbig{}. 
In both cases our experiments showed that PRO outperforms a baseline
classification system, reducing abuse volume by \igreduction
and \fbreduction (respectively) with no negative impact to users. We also
present several case studies that demonstrate how PRO can quickly and
automatically adapt to changes in business constraints, system behavior,
and/or adversarial tactics.
\end{abstract}

%% file: section1_intro.tex
\section{Introduction}\label{sec:intro}

Online Social Networks (\osns) connect billions of people around the world, allowing them to share content, engage in discussions, learn about events and issues, find employment, buy and sell goods, and undertake any number of other useful activities. However, the very utility and popularity of \osns attracts malicious actors who want to take advantage of the \osn's users for their own (usually monetary) gain. As a result, spam~\cite{ceas2010}, phishing~\cite{jagatic2007social}, fake engagement~\cite{dekoven2018following}, abusive accounts~\cite{xu2021deep}, and data scraping~\cite{musa} have become important problems for both academia and the \ifanon \osns \else social networks \fi themselves.

The traditional approach to fighting \osn abuse has been to build {\em binary classifiers} that distinguish ``benign'' content, entities, or actions from ``malicious'' ones. The \osn then runs these classifiers either synchronously or asynchronously and blocks or removes the detected abuse. A great deal of research has gone into designing better and better classifiers (see Section~\ref{sec:related_abuse} for a survey), but even the best known classifier will make thousands to millions of mistakes when run on {\em every} account or action in a large \osn, leading to degraded experiences for the users who encounter these mistakes.

Most binary classifiers produce classifications by comparing a numerical score against a threshold. For instance, scores above the threshold will result in classifying a piece of content as ``abusive.'' Tuning the threshold allows the operator to control the tradeoff between abuse and user experience in a rudimentary way. The \osn can set a ``budget'' on the model's precision (e.g., choose the lowest threshold such that at least $X\%$ of spam is classified correctly) or the false positive rate (e.g., choose the lowest threshold such that at most $Y\%$ of benign content is classified incorrectly). The modeling goal then becomes improving the classifier's recall within these constraints.  However, producing only classification results is insufficient to prevent or mitigate abuse --- eventually, the \osn must take actions with real-world consequences.

Even in a simple setting such as spam detection, where the obvious action is to simply remove the content, there are many factors to consider when taking action:
\begin{itemize}
\squeezelist
    \item Classifiers are never 100\% precise, so users may complain that their benign content is being taken down.
    \item Some users may not be aware of the site's content policy; these users have the potential to be educated and thus improve the quality of their content.
    \item Classifiers have some latency, so malicious users may post abusive content faster than it can be taken down, leading to more spam on the \osn.
\end{itemize}
In the binary-classification approach described above, there is only one parameter we can adjust to balance these competing considerations. It follows that if we want to improve user experience without degrading our ability to remove abusive content, our only avenue is to improve the model, which is a difficult and time-consuming effort.

A first direction of improvement is to {\bf expand the set of possible enforcement actions} beyond the binary ``hard block'' or ``allow.'' For instance, we can add a third, ``soft'' enforcement action: an action that introduces some friction but does not completely block a user. Such actions are less disruptive to benign users while (possibly) being less effective at stopping abuse. We can now segment our classification results into three groups instead of two: the ``worst'' content gets blocked as before, the ``best'' is let through, and the ``suspicious'' is routed through the ``soft'' action. By adjusting the relative sizes of the three groups, we gain an extra degree of freedom to control the tradeoff between amount of abuse blocked and negative impact to user experience.

As a real-world example, almost every website that has a login page chooses to block some logins, let some through, and send the rest through additional checks such as a CAPTCHA or SMS code verification. CAPTCHA and SMS code verification are examples of ``soft'' enforcement actions that lower the cost of false positives, and thus allow the website to send ``suspicious'' users through some non-zero level of enforcement, which reduces false negatives. However, a multi-class classification approach would still run into the problem of defining ``worst,'' ``best,'' and ``suspicious'' in a way that is sufficiently precise to generate objective and low-noise labels.

This additional complexity compels us to consider the action-selection problem from an entirely new perspective: that of {\em adaptive control theory}~\cite{aastrom2008adaptive} and {\em reinforcement learning}~\cite{sutton2018reinforcement}, which are frameworks that focus on decision-making with causal consequences. Reinforcement learning (RL) uses observations from previous actions to choose actions that optimize a reward (in our example, a quantity that captures the amounts of abusive content posted and benign content blocked) while also implementing a data-collection strategy that yields adaptation to non-stationary conditions. Continuous exploration is especially important in an adversarial environment, as future abusive behavior will change in response to our actions on past examples. The full technical details of our formalization appear in Section~\ref{sec:opt}.

In Section~\ref{sec:algo} we present a reinforcement-learning system for fighting abuse on \osns, which we call {\em Predictive Response Optimization} (PRO). We describe the system in terms that apply to {\em any} type of abuse, as long as it can be measured in some way. The system revolves around models that predict the cost and benefit of each action (the {\em contextual multi-armed bandit} setting~\cite{Li_2010}), overlaid with a {\em model predictive control} framework to adjust tradeoffs amongst various abuse and user-experience metrics.

In Section~\ref{sec:system_in_practice} we describe our application of PRO to the problem of reducing scraping activity in \osns. We defined metrics, implemented the system on \fbig, and collected data for two weeks on each. Our experiments showed that in both cases PRO stops significantly more abuse than a baseline system that determines actions by applying a set of hand-coded rules to classifier outputs. Specifically, we {\bf reduced abuse volume by \igreduction on \ig and \fbreduction on fb, with no negative impact on ``benign usage'' metrics}.\footnote{The difference of an order of magnitude between the two results is due to the relative maturity of the experimental and control groups on the two \osns. For details see Section~\ref{ss:results}.}

In Section~\ref{sec:system_in_practice} we also detail five case studies illustrating how PRO can quickly and automatically adapt to changes in business considerations, system behavior, and/or adversarial tactics. Specifically, our experiments show that:
\begin{itemize}
\squeezelist
	\item Adding a new user metric led to an \smsreduction reduction in SMS expenditures with no increase in abuse volume.
	\item After we saw signs of over-enforcement on \surface, adding a new user metric led to a \mwebdaulossreduction decrease in user churn.
	\item Introduction of a new enforcement action reduced abuse volume by \newresponsereduction without any manual intervention.
	\item When a bug changed the behavior of a particular enforcement action, the system adjusted automatically to stop selecting the action.
	\item When adversaries began to evade a particular enforcement action, the system adjusted automatically to select that action less frequently.
\end{itemize}

\myparagraph{In summary, our contributions are:}
\begin{itemize}
\squeezelist
    \item We introduce the perspective that {\em selection of enforcement actions,} rather than binary classification, is the true goal of abuse-fighting systems in online social networks.
    \item We formalize action selection as a {\em reinforcement learning problem} that attempts to balance abuse volume against cost of blocking benign users or content.
    \item We design {\em Predictive Response Optimization} (PRO), a large-scale reinforcement-learning system for action selection and the first application of RL to abuse reduction.
    \item We implement the PRO system on \fbig and observe that it significantly reduces scraping activity with no negative impact to users.
    \item We describe a number of case studies that demonstrate the ability of PRO to adapt to changing conditions with minimal manual intervention.
\end{itemize}

%% file: section2_background.tex
\section{Background and related work}\label{sec:background}

Abuse on social-networking platforms can take various forms. Attackers can create fake accounts (or ``sybils'')~\cite{xiao2015detecting,gong2014sybilbelief,danezis2009sybilinfer,clickstream,uncovering_social} or compromise existing accounts, which they can then use to spread phishing links~\cite{gao2010detecting, grier2010spam}, post fraudulent reviews and advertisements~\cite{al2021spam}, disseminate fake news~\cite{elazab2018fraud}, or make fraudulent payments~\cite{caldeira2014fraud}. Such attacks motivate research on detecting and removing abusive content on social networks. 

In other types of attacks, attackers collect user data from social networks to use later for malicious purposes. Previous studies demonstrate how attackers can scrape social networks to collect user information that is then used for targeted spam and phishing~\cite{balduzzi2010abusing,jagatic2007social,brown2008social}. These attacks motivate research on detecting and blocking automated activity (i.e., ``bots'') on social networks~\cite{orabi2020detection}.

\subsection{Related work in abuse detection}
\label{sec:related_abuse}

Most prior works focus on how to detect abusive content and/or users by leveraging machine learning techniques that differentiate abusive from non-abusive entities. Such research often aims to improve the precision or recall of various supervised ML algorithms such as $k$-nearest neighbors, random forests, naive Bayes, decision trees, and neural networks~\cite{al2021spam,elazab2018fraud,caldeira2014fraud,acm2010,ceas2010,LCAI2011}. Given labeled data, these algorithms can use network information, behavioral patterns, or features generated from the content itself to detect adversaries~\cite{IEEE2016} and even to continuously detect adversaries who try to elude the model~\cite{KDD2004}. One challenge in using supervised methods is the reliance upon labeled data, which can limit the scalability of these approaches~\cite{IEEE2016}. This limitation, combined with the fact that abusive entities such as spammers also tend to act collusively, has led to the development of unsupervised methods, including graph-based and clustering methods \cite{WWW2013,ACM2014,WWW2016,IEEE2016}. Whether supervised or unsupervised, these approaches focus on detecting abusive entities but ignore the negative impact of incorrect detection. Such negative impact arises only after an enforcement action is applied to the detected entity.

\subsection{Enforcement methods}\label{sec:bg_enforcements}

After identifying abusive entities, anti-abuse systems apply {\em enforcement actions} to induce the potentially abusive actor to stop or change their behavior. In the previously discussed abuse-detection systems, generally a single action is applied for all detected abusive entities (e.g., deleting the detected fake profile or abusive content~\cite{acm2010}). In some cases, the system uses the detection confidence to choose between a ``mild'' enforcement (e.g., removing fake engagement) or a ``harsh'' enforcement (e.g., taking down accounts)~\cite{WWW2016}.

Some \osns, such as YouTube~\cite{youtube} and Facebook~\cite{meta}, use a ``strike'' system to carry out an escalating series of enforcement actions to prevent repeat offenders. On both YouTube and Facebook, the process begins with a warning. On YouTube, each strike temporarily restricts content creation, and after 3 strikes in a 90-day period the offending channel is removed~\cite{youtube}.
On Facebook, strikes 2--6 yield temporary feature-specific restrictions, while additional strikes trigger content-creation restrictions of increasing length~\cite{meta}. 
Currently there is no literature on abuse-minimization systems that addresses the problem of choosing between multiple enforcement methods based on a set of constraints; our work fills this gap.

\subsection{CAPTCHAs, challenges, and verification}

A widely used technique for combatting abuse that leverages automation (such as fake engagement or data scraping) is to present ``challenges'' that are difficult for bots to solve while remaining relatively accessible for humans. Von Ahn et al.~\cite{von2003captcha} introduced several practical proposals for designing ``CAPTCHA'' schemes for this purpose: problems that most humans can solve easily but computer programs cannot. A recent survey~\cite{guerar2021gotta} identified 10 types: text-based, image-based, audio-based, video-based, math-based, slider-based, game-based, behavior-based, sensor-based, and CAPTCHAs for liveness detection. CAPTCHAs have been used to ensure the safety of network applications~\cite{banday2011study, longe2009checking}, including chat rooms, email registration, online auctions, file sharing, and polls~\cite{pope2005human}.

Many systems also commonly use {\em multi-factor authentication} (MFA) to verify that an entity (e.g., a web session or social-media account) is being controlled by an authorized owner and not a malicious actor. Ometov et al. provided a survey of MFA methods~\cite{mfa}. Some common strategies include password re-verification, hardware tokens, voice biometrics, facial recognition, and phone or email verification. Both CAPTCHA and MFA actions are included as enforcement options within our system.

\subsection{Optimization and control}

One of our core contributions is to apply optimization and control methods to select the ``best'' enforcement action. Such methods have been leveraged in many application domains, but to date this list does not include \osn abuse.

{\em Model Predictive Control} (MPC)~\cite{arroyo2022reinforced,hewing2020learning} is a control approach that leverages a system model to predict or simulate how different inputs to the system will affect the system's output up to a certain time in the future. Based on these predictions, the operator can choose the  input that leads to most desirable output. They can then repeat the control process with new observations~\cite{holkar2010overview,hewing2020learning,recht2019tour}. MPC is robust, sample-efficient, and able to handle enforcing constraints, but it can be difficult to apply to complex systems (e.g., constructing a system model that also models uncertainty)~\cite{arroyo2022reinforced}. 

{\em Reinforcement learning (RL)}~\cite{sutton2018reinforcement} is an alternative control approach that can predict outputs of complex systems without a pre-defined model. Technically, RL is a class of algorithms that maximize specified objectives by learning from prior actions. RL algorithms include ``model-free'' approaches such as Q-learning, which learn the ``quality'' of a particular action executed while the system is in a particular state~\cite{watkins1992q,mnih2015human,kalweit2020deep}. ``Model-based'' approaches, on the other hand, learn a model for the system~\cite{sutton2018reinforcement,recht2019tour} and are related to ``system identification'' from the control field~\cite{Ljung99,recht2019tour}. RL has been used in applications such as board games (Go, chess), arcade games (PAC-MAN)~\cite{jansen2020safe, silver2018general}, recommender systems ~\cite{chow2017risk, zhao2021dear}, transport scheduling~\cite{basso2022dynamic, chinchali2018cellular, singh2020hierarchical, li2019constrained}, finance~\cite{abe2010optimizing, krokhmal2002portfolio}, and autonomous driving~\cite{gu2022constrained, isele2018safe, krasowski2020safe, kalweit2020deep}.

Prior research has combined MPC with RL to add constraints and improve safety of the RL system~\cite{arroyo2022reinforced,zanon2020safe}. Our work follows this approach, leveraging RL for learning the system and taking actions at an entity level and leveraging MPC to apply global constraints.

In the field of security, RL-based solutions have been proposed to defend against cyber-attacks on various IoT systems~\cite{uprety2020reinforcement}. RL has also been used to improve security in cognitive radio networks~\cite{ling2015application} and to detect botnets~\cite{alauthman2020efficient}. MPC has also been applied within a security context to detect cyber-attacks in microgrids~\cite{habibi2021secure}. Other work focuses on detecting attacks within networked systems controlled with MPC~\cite{barboni2018model}. However, to our knowledge, there is no prior work in any security context that explores the application of machine learning to optimize action selection from a list of possible enforcement actions.

%% file: section3_problem.tex
\section{Abuse minimization as a constrained optimization problem}\label{sec:opt}

In this section we reframe the goal of anti-abuse systems in online social networks: rather than existing to {\em classify} content or behavior, such systems exist to {\em optimize the tradeoff} between abuse reduction (e.g., removing spam) and impact on user experience (e.g., too much benign content removed). Specifically, our position is that {\bf anti-abuse systems are solving the {\em constrained optimization problem} of minimizing abuse volume within a ``budget'' of operational costs.} We conjecture that this reframing will enable us to devise a system that blocks more abuse than a classification system, without adversely impacting user experience.

\subsection{Abuse minimization is a tradeoff problem} 
\label{sec:business_constraints}

Anti-abuse systems seek to achieve the dual aims of both reducing abuse prevalence and minimizing costs that arise as a result of practical considerations. Some of these quantifiable business considerations are:

\begin{itemize}
\setlength\itemsep{-2pt}
    \item Size of the operational team that processes appeals of incorrect content deletions or account disables.
    \item Users choosing to leave the platform because of enforcement actions against their content.
    \item Negative impact to usability/engagement metrics on the \osn due to increased user friction (e.g., challenges and verifications).
    \item Computer and network hardware needed to run the anti-abuse system itself.
\end{itemize}

\begin{figure}[htbp]
\centering
\captionsetup{justification=centering,margin=1cm}
\includegraphics[width=8cm]{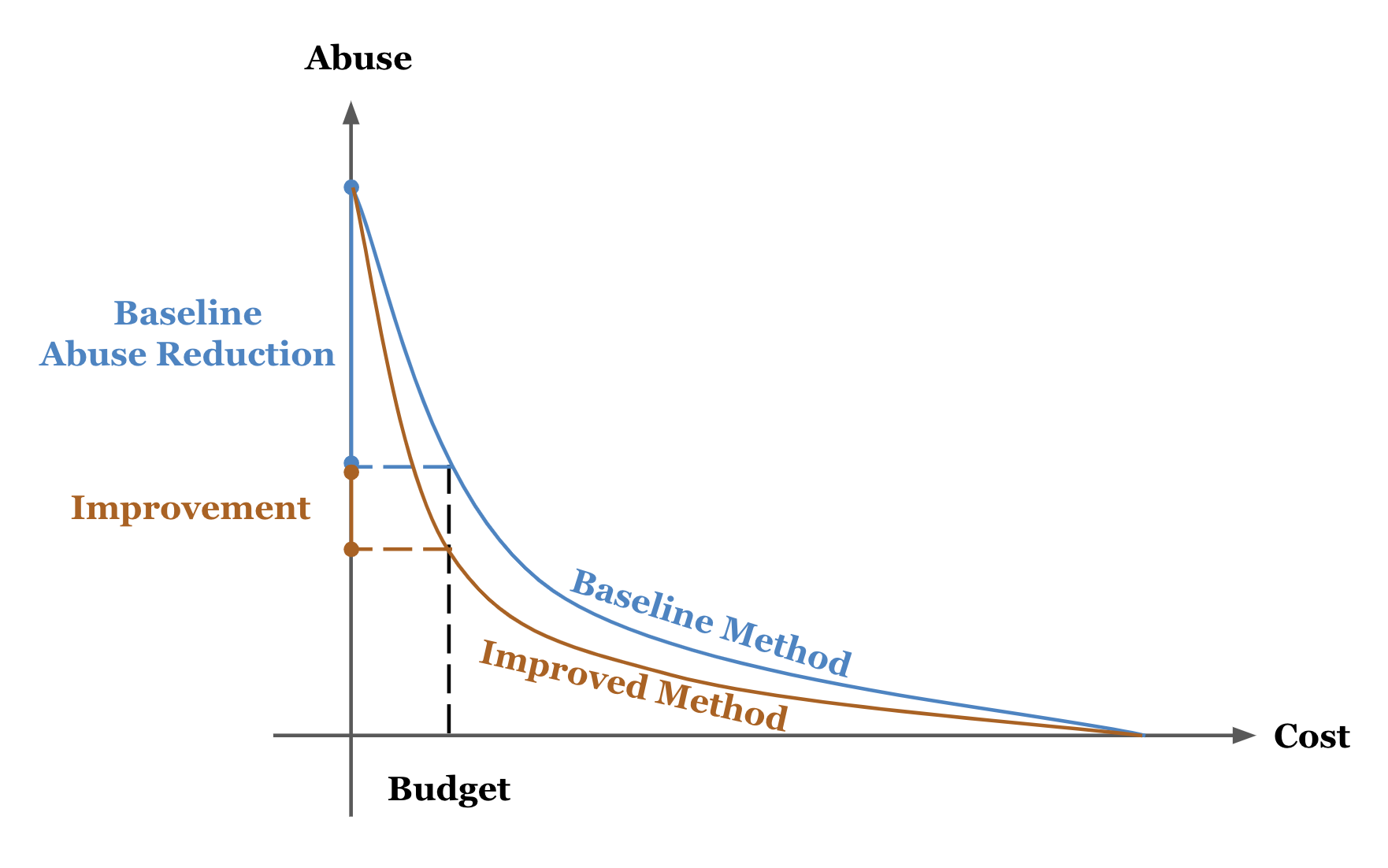}
\caption{Two conceptual anti-abuse methods tuned under constraints.}\label{fig:pareto}
\end{figure}

Conceptually, the abuse-cost tradeoff can be visualized as a Pareto frontier~\cite{Wilkinson_2006} if we simplify and reduce operational cost considerations to a single ``cost'' dimension (Figure~\ref{fig:pareto}).
As illustrated, abuse can be fully eradicated by any method if the cost axis is not constrained --- after all, if we block {\em all} users from using the platform, there will be no abuse left! However, for any anti-abuse method to work in a real-world setting, the system must be tuned to operate under some cost constraints (``budget'' in Figure~\ref{fig:pareto}).
The abuse/cost tradeoff space is also often multi-dimensional; i.e., multiple business-metric constraints can exist simultaneously. 

\subsection{Enforcement actions really matter}\label{sec:enforcement_actions}

On its surface, the ``what to do'' problem seems fairly straightforward to solve: if an entity is abusive, remove it from the \osn! However, if we take content moderation as an example of a typical use case, we see that a policy of ``delete all content classified as abusive'' quickly runs into issues such as those mentioned in the Introduction.

To address these issues while remaining effective at stopping abuse, anti-abuse systems typically employ multiple enforcement actions that are designed to hinder or discourage abusive behavior to varying degrees. At the same time, these actions affect benign (i.e., non-violating) users to different extents, ranging from mild annoyance to total loss of access to their account and the value associated with it.
As an example, in the case of content moderation, a possible set of enforcement actions could include one or more of the following (ranging from least to most intrusive):
\begin{itemize}
\vspace{-2pt}
\squeezelist
    \item No action,
    \item Incrementing a ``strike counter'' for the author~\cite{youtube},
    \item Showing a warning notification to the author,
    \item Down-ranking/reducing visibility of the content,
    \item Asking the author to perform an interactive challenge (e.g., CAPTCHA, MFA) before posting content~\cite{von2003captcha,searles2023empirical},
    \item Deleting the content,
    \item Temporarily banning the author from sharing content,
    \item Temporarily banning the author from the service,
    \item Permanently revoking the author’s access to the service.
\end{itemize}

In ``traditional'' systems, action selection is implemented using a set of hand-written logical rules that incorporate information such as classifier confidence, past enforcement/strike counts, and other features related to the entity. The focus of prior research on the classification problem, while leaving enforcement selection to rule- or strike-based logic, results in three major pitfalls when applied in practical settings:
\begin{itemize}
\squeezelist
    \item Abuse prevalence could remain high due to ineffective enforcement; e.g., as a result of poor optimization against practical constraints.
    \item Fixed action-selection rule logic does not adapt to changing adversarial behavior; e.g., learning to bypass certain enforcement actions.
   \item Introducing new actions requires writing new routing rules and could cause constraint metrics to shift in the wrong direction.
\end{itemize}

Instead of relying on hand-written rules, we propose formalizing the enforcement-selection problem as a constrained optimization problem and then develop a reinforcement-learning algorithm to solve it.

\subsection{Formalizing the optimization problem}\label{sec:problem_math}

Anti-abuse systems evaluate and execute actions over a potentially very large set $\mathcal{E}$ of entities (\osn accounts, pieces of content, IP addresses, and so on). If the system takes action on entity $e \in \mathcal{E}$ at time $t_0$, we can measure the impact of this action on entity $e$ over the subsequent time period $\tau = [t_0, t_1]$ using a set of ``abuse metrics'' $Abuse_{j}$ ($j  = 1, \dots, N_A$) and ``cost metrics'' $Cost_{j}$ ($j = 1,  \dots, N_C$). 

To make our discussion more concrete, we will use the running example of \osn accounts posting spam, with abuse metric $Abuse^*$ being ``number of spam posts during the time period $\tau$'' and cost metric $Cost^*$ being ``number of non-spam posts blocked during the time period $\tau$.'' 

Let $\mathcal{A}$ denote the (finite) set of all possible actions. We define the vector of all ``next actions'' for all entities as  
\begin{equation}\label{eq:action_set}
\mathbf{a} \in \AAA = \left\{ \left(a_e\right)_{ e\in\mathcal{E}}:  a_e \in \mathcal{A} \right\} .
\end{equation}
The set $\mathcal{A}$ contains the special action $A_0$ denoting ``no action,'' plus other actions such as those described in Section~\ref{sec:enforcement_actions}. In our spam example we will let $\mathcal{A}$ be a set of three actions: $A_0 = $ ``no action''; $A_1 = $ ``show a CAPTCHA''; $A_2 = $ ``disable the account.''

At optimization time $t_0$, our goal is to determine the action vector $\mathbf{a}$ that minimizes the total abuse over the period $\tau$
\begin{equation}\label{eq:objective} 
\min_\mathbf{a} \sum_{e \in \mathcal{E}} \sum_{j=1}^{N_A} Abuse_{j}(e \mid \mathbf{a})
\end{equation}
while constraining the cost over the same period
\begin{equation}
	\label{eq:constraints}
	\sum_{e \in \mathcal{E}} Cost_{j}(e \mid \mathbf{a}) \leq B_{j}, \qquad j = 1, \ldots, N_C
\end{equation}
where $B_j \geq 0$ is a global ``budget'' for the $j$th metric. In our concrete example, we want to minimize total spam posts while keeping the total number of blocked non-spam posts below a certain fixed number $B^*$ determined by the business.

We assume that all abuse and cost metrics are normalized in such a way that 
\begin{equation}
    \label{eq:normalization}
Abuse_{j}(e \mid \mathbf{a}^0) = Cost_{j}(e \mid \mathbf{a}^0) = 0,
\end{equation}
where the baseline action vector $\mathbf{a}^0 = \left(A_0, \ldots, A_0\right)$ corresponds to applying ``no action'' to all entities, effectively ``turning off'' the anti-abuse system. Furthermore, we assume that the metrics are signed such that smaller values are ``better'' (i.e., we want to {\em minimize} both abuse and cost). This normalization ensures that $\mathbf{a}^0$ always satisfies the constraints, making the optimization problem feasible (i.e., a solution always exists). 

Let us consider the effect of this normalization on our spam example. An account $e_{spam}$ that posts only spam will (presumably) have 
$Abuse^*(e_{spam} \mid \mathbf{a}) \le 0$
for all $\mathbf{a}$, since (presumably) any nontrivial action will reduce the amount of spam posted by that account. The account will also have $Cost^*(e_{spam} \mid \mathbf{a}) = 0$ for all $\mathbf{a}$ since there are no non-spam posts to block. On the other hand, an account $e_{benign}$ that posts no spam will have $Abuse^*(e_{benign} \mid \mathbf{a}) = 0$ for all $\mathbf{a}$ and 
$Cost^*(e_{benign} \mid \mathbf{a}) \ge 0 $
for all $\mathbf{a}$ since (presumably) any nontrivial action will only decrease the number of non-spam posts; i.e., contribute a non-negative number of blocked posts.\footnote{Note that ``blocked posts'' in this example includes not only non-spam posts blocked directly but also those ``prevented'' relative to the no-action baseline. For example, if the action is to disable the account then the $Cost^*$ metric attempts to estimate how many non-spam posts the user ``would have made'' had they not been blocked.}

In practice, at optimization time we don't have access to any of the abuse and cost metrics since they refer to a future time period and will only be available after a time delay.  Moreover,  the constrained optimization problem (\ref{eq:objective})--(\ref{eq:constraints}) couples together all the entities, which makes it unfeasible to solve at high frequency. Therefore, we also consider an unconstrained relaxation consisting of maximizing the following ``reward'' function with respect to $\mathbf{a}$:
\begin{equation}
\label{eq:reward}
r(\mathbf{x},\mathbf{a}) = -\sum_{e \in \mathcal{E}} r_e(\mathbf{x},\mathbf{a}), \quad r_e(\mathbf{x},\mathbf{a}) = \sum_{j=1}^N w_j \cdot m_{j}(e \mid \mathbf{x}, \mathbf{a})
\end{equation}
\noindent where we have combined together all the metric functions
\[
(m_{1}, \ldots, m_{N})= (Abuse_{1}, \ldots, Abuse_{N_A}, Cost_{1}, \ldots, Cost_{N_C}),
\]
(setting $N = N_A + N_C$) and introduced multipliers $w_j\geq0$ that determine the relative weighting of each $Abuse$ and $Cost$ metric. The $w_j$ also implicitly convert all metrics into a common unit; for example an abuse metric might be in units of spam posts while a cost metric might be in units of benign users blocked.  The quantity 
\begin{equation}\label{eq:state}
\mathbf{x} = \left({x}_e \in \mathcal{X}\right)_{e\in\mathcal{E}}
\end{equation}
represents the state information (features) available at optimization time for all entities, where $\mathcal{X}$ is the ``feature space'' used to describe the state for a particular entity. This information allows us to leverage predictive models obtained via machine learning to approximate a solution to the optimization problem. Applying the notation in \eqref{eq:reward} to our spam example gives us the per-entity reward function
\begin{equation}
    \label{eq:example_reward}
 r_e(\mathbf{x},\mathbf{a}) = Abuse^*(e \mid \mathbf{x},\mathbf{a}) + w \cdot Cost^*(e \mid \mathbf{x}, \mathbf{a}).
 \end{equation}
Here we can interpret $w$ as the ``relative weight'' of the two harms being traded off: blocking one non-spam post is equivalent to allowing $w$ spam posts.

In general, maximizing (\ref{eq:reward})  is not equivalent to solving (\ref{eq:objective})--(\ref{eq:constraints}), though in some cases there exist Lagrange multipliers $w_j$ that make the two problems exactly equivalent. Nevertheless, the multipliers $w_j$ can be adjusted periodically to ensure that the optimal solution of (\ref{eq:reward}) tracks the optimal solution of (\ref{eq:objective})--(\ref{eq:constraints}) as closely as possible.  We show in the next section that the unconstrained relaxation (\ref{eq:reward}), combined with suitable modeling assumptions, enables high-frequency decision making by decoupling the optimization across entities.

%% file: section4_rl.tex
\section{Solving the optimization problem}\label{sec:algo}

\begin{figure*} %
\centering
\captionsetup{justification=centering,margin=0.4cm}
\includegraphics[width=0.75\textwidth]{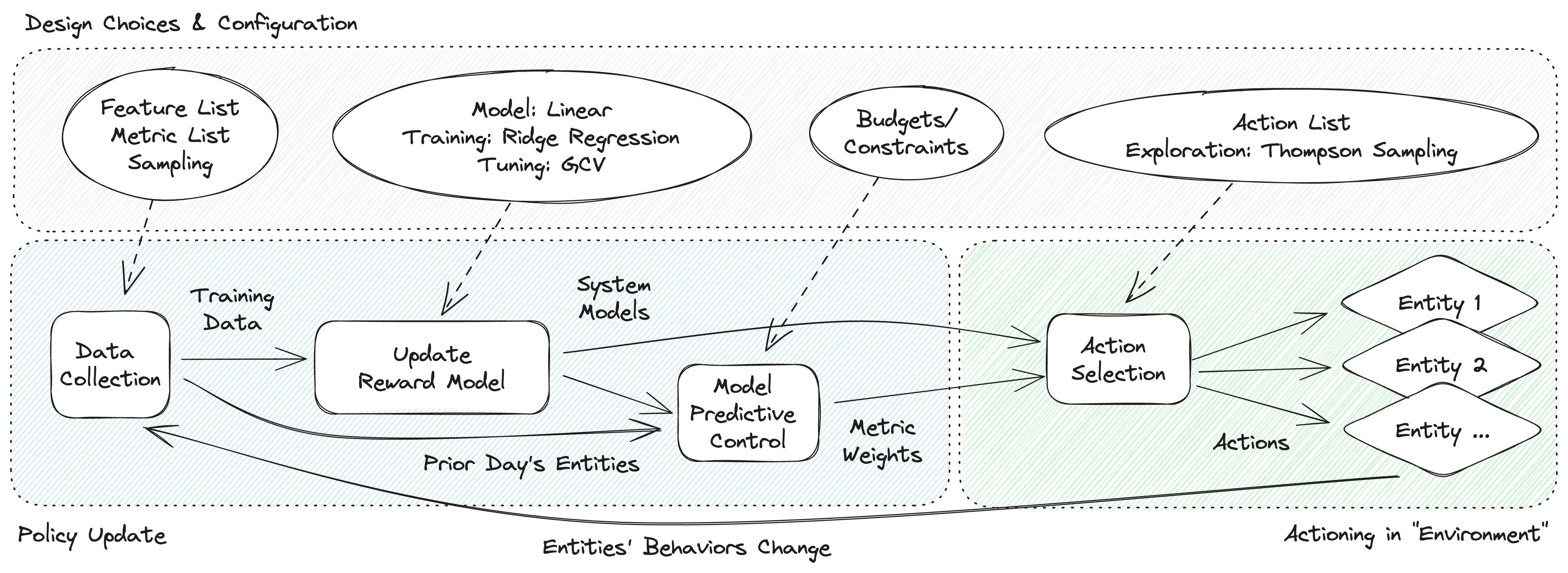}
\caption{Predictive Response Optimization system design}\label{fig:architecture}
\end{figure*}

In this section we describe a strategy to approximate solutions to the optimization problem (\ref{eq:objective})-(\ref{eq:constraints}). Our strategy combines elements of reinforcement learning (RL) with model predictive control (MPC) and consists of two components:
\begin{enumerate}
\squeezelist
	\item Optimizing actions $\mathbf{a}$ at entity level,
	\item Optimizing multipliers $w$ to enforce global constraints.
\end{enumerate}
Figure~\ref{fig:architecture} provides an overview of the system components and the design choices for each component. 

We will show that entity-level action optimization can be described as a \emph{contextual multi-armed bandit problem}. By introducing suitable modeling assumptions, we are able to decide actions asynchronously for each entity, at arbitrary time intervals. Similarly, our approach to finding optimal multipliers can be seen as a form of stochastic \emph{model predictive control} (MPC) aimed at enforcing global constraints. Other works combining RL with MPC include ~\cite{arroyo2022reinforced,zanon2020safe}. 
 
\subsection{Optimizing action selection}\label{sec:model}

In  reinforcement learning (RL) terminology, the optimization system (a.k.a.~\textit{agent}) acts within an {\em environment}. The agent takes an {\em action} chosen from a set of possible actions depending on the agent's {\em state} and then receives an application-specific {\em reward}. The choice of action is based on the agent's {\em policy}, which in addition to selecting actions to maximize the expected reward (``exploit''), also strives to gain information (``explore'') to improve the policy itself ~\cite{sutton2018reinforcement}.

Maximization of (\ref{eq:reward}) can be readily framed as an RL problem by defining the state $\mathbf{x}$ as in (\ref{eq:state}) and the set of actions $\mathcal{A}$ as in (\ref{eq:action_set}). Equation (\ref{eq:reward}) defines the reward as a weighted sum of the values contributed by each entity towards each metric (in our example, the amount of spam content and number of erroneous deletions). Each of these metrics is a cumulative quantity aggregating data with timestamps between the time $t_0$ when an action is chosen and a future time $t_1$. We call the interval $\tau = [t_0, t_1]$ the {\em time horizon}. The environment is modeled via the state and the metric functions.

After applying an action, an entity's behavior may change, altering the received reward (i.e., metric values) over the time horizon $\tau$. In our example, if we temporarily ban an account $e$ from sharing content, then $Abuse_1(e)$ counting number of spam posts may decrease but $Cost_1(e)$ counting number of non-spam posts blocked may increase (relative to their values if no action were taken). 

In more general RL problems, the reward is a function of transitioning from state $\mathbf{x}$ to state $\mathbf{x}'$ via action $\mathbf{a}$: $r(\mathbf{x},\mathbf{a},\mathbf{x}')$~\cite{sutton2018reinforcement}; however in Equation~\eqref{eq:reward} we simply have $r(\mathbf{x},\mathbf{a},\mathbf{x}') = r(\mathbf{x},\mathbf{a})$ (i.e., the \textit{contextual multi-armed bandit} setting \cite{Li_2010}).  In other words, instead of modeling state transitions from multiple agent actions as a Markov decision process~\cite{puterman2014markov}, we aim to predict the incremental reward from each individual action, thus simplifying the RL problem.

The goal of the RL problem is to maximize the cumulative reward over multiple action-selection events (also known as the \textit{return}). Without loss of generality, action selection can be viewed as sampling $\mathbf{a}$ from a probability distribution $\Pi(A \mid \mathbf{x})$ conditioned on the state $\mathbf{x}$; here $A$ is a random variable taking values in $\AAA$ and we call the distribution $\Pi$ the \emph{policy}~\cite{sutton2018reinforcement}. (Note that this framing includes the case when actions are chosen deterministically.) 

Our approach to maximizing cumulative reward is to first build a predictive model $\pi(R \mid \mathbf{x},\mathbf{a})$ describing the probability distribution for the next reward value (\ref{eq:reward}) modeled as a random variable $R$ conditioned on a given pair of states and actions $(\mathbf{x},\mathbf{a}) \in \mathcal{X} \times \AAA$. We then implicitly define our policy $\Pi$ by its sampling mechanism:
\begin{equation}\label{eq:general_policy}
\mathbf{a} \gets \Pi(A \mid \mathbf{x}) := \argmax_{\mathbf{z} \in \AAA} \Big( r \gets \pi(R \mid \mathbf{x}, \mathbf{z}) \Big)
\end{equation}

In our example of spam posts, we realize the policy in \eqref{eq:general_policy} by building a predictive model for the reward function \eqref{eq:example_reward}, predicting the reward for each component of the action vector $\mathbf{z} \in \AAA$, and setting $\mathbf{a}$ to be the action vector $\mathbf{z}$ with the greatest reward.

Our sampling approach is derived from {\em Thompson sampling}~\cite{thompson1933likelihood,agrawal2012analysis}; in particular, the variability inherent in the model $\pi$ allows us to balance exploration and exploitation (a classic challenge in reinforcement learning) by tuning model parameters~\cite{sutton2018reinforcement}.  Exploration is necessary to improve the model's understanding of how different actions impact each metric in various regions of the feature space. It is particularly important in adversarial environments, as malicious actors may learn to work around certain enforcement actions, rendering a previously heavily exploited enforcement action ineffective (see Section~\ref{ss:adaptation} for an example). Exploration is also useful as a mechanism to onboard newly developed enforcement actions to the system (see Section~\ref{ss:new_action} for an example). 

\subsection{Reward models}\label{sec:reward_prediction}

In general, taking action on one entity may affect other entities' metrics. For instance, when banning a user account for spamming, other users will not be exposed to the spam anymore. However, modeling all possible interactions is prohibitively expensive. Therefore, in the model described in this section we neglect the impact of actions taken on a given entity to metrics for other entities.  This decoupling allows us to determine actions in real time whenever a decision for a particular entity is needed. Formally, we set
\begin{equation} \label{eq:decoupling_entities}
\pi (R\mid \mathbf{x},\mathbf{a}) = \prod_{e \in \mathcal{E}} \hat \pi (R_e\mid {x}_e,a_e),
\end{equation}
where $R_e$ is a random variable modeling the portion of the reward contributed by entity $e$, the contextual features $x_e$ summarize all information about entity $e$ that is deemed useful to predict future metric values, and $\hat \pi$ is a (global) model that predicts rewards at the entity level given the contextual features $x_e$ and the action $a_e$ applied to $e$. 
Examples of features relevant for abuse minimization at user-account level include:
\begin{itemize}
\squeezelist
    \item Account properties such as age, classifier scores\ifanon\else~\cite{xu2021deep}\fi, number of sessions, or frequency of requests;
    \item Time series of metric values for the user;
    \item History of prior actions taken on the account (e.g., a vector of whether each action was taken the day before, 2 days before, and so on);
    \item Classifier scores for the account's content (e.g., probability of spam);
    \item IP statistics such as number of requests in a time window or number of accounts using the IP;
    \item Request features such as requested URL or browser type.
\end{itemize}

Let $\mathcal{T}$ be the \emph{training window}; i.e., a set of timestamps in the past for which we have entity-level historical data. For each $t \in \mathcal{T}$, let $x_e^t$ be the value of the feature vector (i.e., state) $x_e$ at time $t$, let $a_e^t$ be the action taken on entity $e$ at time $t$, and let $m_j^t$ be the value of the metric $m_j$ starting at time $t$ (i.e., over the period $[t, t+t_1-t_0]$). For each metric $m_j$ and action $A_k \in \mathcal{A}$, we construct datasets
\[
\mathcal{D}_{jk} = \left\{\left(x_e^t, m^t_{j}(e \mid x_e^t, a_e^t)\right) \quad \forall e \in \mathcal{E},\ t \in \mathcal{T} : a_e^t = A_k\right\}
\]
that document historical (state, metric-value) pairs for each action and metric. In our spam example, $\mathcal{D}_{1k}$ is the dataset consisting of all (feature-vector, $Abuse^*$-value) pairs for the entities $e$ that received action $A_k$ at some time during the training window, while $\mathcal{D}_{2k}$ is the corresponding set of (feature-vector, $Cost^*$-value) pairs.

In our implementation we sample $t \gets \mathcal{T}$ in such a way that the amount of data decreases exponentially as the timestamp $t \in \mathcal{T}$ goes back in time, biasing the dataset towards recent information while still keeping a fraction of older data. When first deploying the system (the ``cold-start problem''), these datasets can be initialized with data collected from a baseline rule-based system. Subsequently, they consist of data collected after applying actions chosen by the RL system.

After constructing the datasets, we model the metric values $m^t_{j}$ as noisy Gaussian samples
\[
m^t_{j}(e \mid x_e^t, a_e^t) \gets \mathcal{N}( \nu_j(x_e^t,a_e^t), \epsilon) 
\]
where $\nu_j$ ($j  = 1, \ldots, N$) are realizations of independent Gaussian Processes (see \cite{williams2006gaussian}) with mean zero and kernel (covariance function)
\begin{equation}\label{eq:kernel}
K_j((x_1,a_1),(x_2,a_2)) = (\phi_j(x_1)^T \phi_j(x_2))\cdot \delta(a_1,a_2).
\end{equation}
Here the maps $\phi_j: \mathcal{X}\rightarrow \mathbb{R}^D$ represent ``feature transformations'': functions that process the raw input features, extract interactions, and output $D$-dimensional numerical vectors. For example, $\phi_j$ might include taking $\log$ of account age to reduce skew, applying one-hot encoding to categorical features~\cite{one-hot}, or combining action history data to create a summary feature ``number of times blocked in the last 14 days.'' The function $\delta(x,y)$ is 1 if $x = y$ and 0 otherwise.

Given the modeling assumptions above, we can use Gaussian Process Regression~\cite{williams1998prediction} to obtain predictive distributions $\nu_j$ for all metrics $m_j$. In view of our assumption that the metrics appearing in Equation~\eqref{eq:reward} are independent, we obtain reward models of the form
\begin{equation}\label{eq:model_structure}
	\hat \pi (R_e \mid x_e, a_e) = \mathcal{N}\left(\sum_{j=1}^N w_j\cdot \mu_{j}(x_{e},a_e),\sum_{j=1}^N w_j^2 \cdot \sigma_{j}^2(x_{e},a_e) \right).
\end{equation}
Moreover, in view of the kernel structure, we have
\begin{equation}
\label{eq:decoupling_actions}
	\mu_{j}(x_{e},a_e) = \mu_{jk}(x_e), \quad  \sigma_{j}^2(x_{e},a_e) = \sigma_{jk}^2(x_{e}) 
\end{equation}
\noindent where $k$ is the index in $0, \ldots, \abs{\mathcal{A}}-1$ such that $a_e  = A_{k} \in \mathcal{A}$, and the functions $\mu_{jk}$ and $\sigma^2_{jk}$ are defined as
\begin{equation}\label{eq:inference}
\mu_{jk}(x) = \phi_j(x)^T \theta_{jk}, \quad \sigma_{jk}^2(x)  =  \epsilon  \cdot \left( \phi_j(x)^T  \Sigma_{jk} \phi_j(x) \right).
\end{equation}
The parameters $\theta_{jk}$ and $\Sigma_{jk}$ can be learned by pooling historical data across all entities to train $N \cdot \abs{\mathcal{A}}$ independent models (one for each metric $m_j$ and action type $k$). Specifically, we can compute
\begin{align}
\label{eq:compute_sigma} \Sigma_{jk} & =  \left(X_{jk}^T X_{jk} + \lambda_{jk} \cdot I_D\right)^{-1} \quad  \in \mathbb{R}^{D \times D}, \\
\label{eq:compute_theta} \theta_{jk} & =  \Sigma_{jk} \cdot X_{jk}^T \cdot Y_{jk} \quad \in \mathbb{R}^D, 
\end{align}
where $X_{jk}$ is a $\abs{\mathcal{D}_{jk}} \times D$ matrix with rows $\phi_j(x_e^t)$ for all $(e,t)$ in the dataset $\mathcal{D}_{jk}$, $Y_{jk}$ is a $\abs{\mathcal{D}_{jk}}$-dimensional column vector containing the corresponding metric values $m_{j}^t$, and $I_D$ is the $D \times D$ identity matrix. We set Ridge regularization parameters $\lambda_{jk}$ by optimizing generalized cross-validation scores~\cite{golub1979generalized}, while the noise variance $\epsilon$ is a global parameter that can be used to control the rate of exploration. (In our experiments we set $\epsilon = 0.05$.) In our spam example with two metrics and three actions, this process gives us 6 predictive models, each described by $D^2+D$ parameters $\Sigma_{jk}$, $\theta_{jk}$.

\myparagraph{Computational complexity.} For the training step, since the parameter computations~\eqref{eq:compute_sigma}--\eqref{eq:compute_theta} are linear, they can still be feasible for datasets $\mathcal{D}_{jk}$ with millions of entries and feature dimensionality $D$ in the hundreds. In particular, the limiting step is the matrix multiplication and inversion of \eqref{eq:compute_sigma}, which is $O\left(D^2 (D + \abs{\mathcal{D}_{jk}})\right)$ in our implementation.

For the inference step, the simplicity of the models described in \eqref{eq:model_structure}--\eqref{eq:inference} (where the limiting step is $O(D^2)$) enables us to process a very large volume of requests at inference time, interpret model weights, and understand the directional impact of features on predictions and decisions. However, we find in practice that we must frequently update the model because the highly non-stationary nature of both the adversarial and business landscapes results in constant shifts in both cost budgets and system effectiveness. (See Sections~\ref{ss:business} and~\ref{ss:adaptation} for examples.) 

\subsection{Enforcing business constraints} \label{sec:mpc}
The multipliers $w_{j}$ (Equation~\eqref{eq:reward}) control the tradeoffs amongst the various abuse and cost metrics. While reinforce-ment-learning approaches often leverage a single model that predicts a ``quality estimate'' of the state resulting from a particular action applied to a given state~\cite{mnih2015human}, by instead decomposing the reward into a weighted combination of individual metric models we can adjust these metric tradeoff weights ``on the fly'' without model retraining. This property allows us to make quick adjustments if we observe the system being ``too harsh'' (i.e., cost constraints are violated) or ``too lenient'' (i.e., we are not using our entire cost budget). We adjust the multipliers automatically using a controller 
designed to maximize the estimated reward aggregated over all entities and  over a future time period, while attempting to satisfy the budget constraints for the system (Section~\ref{sec:business_constraints}).  Our approach can be interpreted as an instance of stochastic Model Predictive Control (MPC)~\cite{mesbah2016stochastic} with multiple control variables. 

In MPC, a model of the system or ``plant'' to be controlled (in our case the RL ``environment'') may be specified via state space or a transfer function, or learned via means such as system identification. Then, using this plant model, MPC will predict the plant's outputs for various controller outputs. These predictions can be at multiple time horizons, e.g., +1 second, +1 day, etc. MPC will select the controller outputs that are predicted to yield values closest to the plant's desired output. Additional constraints can also be applied, e.g., limiting the plant outputs to a certain range. 

The MPC framework is well suited to our goal of minimizing abuse while enforcing the business constraints for our system. As described previously, we continually learn and update our reward models. Using these models, we can leverage the prior day's data to predict the overall action and metric distributions for a variety of metric tradeoff weights, i.e., the ``simulation'' in MPC. This process is a form of multi-variable MPC where the control horizon is one step ahead. Then, if we set the metric tradeoff weights $w_j$ to be the parameters controlled by MPC, the controller can pick these parameters such that the constraints are met (in expectation). Note that using the prior day's data for simulation assumes the distribution of features and metrics does not shift substantially from one day to the next.

Consider again the optimization problem described in Equations~(\ref{eq:objective})--(\ref{eq:constraints}), now assuming that the action vector $\mathbf{a}$ is obtained by sampling from the policy $\Pi$ defined in (\ref{eq:general_policy}).  Since $\Pi$ depends on the multipliers $w_j$ via (\ref{eq:decoupling_entities}) and (\ref{eq:model_structure}), we can now consider optimizing the objective (in expectation) with respect to $w$: 
\begin{align*}
\min_w \sum_{e \in \mathcal{E}} \sum_{j=1}^{N_A} \mathbb{E}\left[w_j \cdot m_{j}(e \mid \mathbf{a})\right], & \quad \mbox{subject to} \\
\sum_{e \in \mathcal{E}} \mathbb{E} \left[w_j \cdot m_{j}(e \mid \mathbf{a})\right] \leq Budget_{j}, & \quad j = N_A+1, \ldots, N.
\end{align*}
We are now left with optimizing $N$ weights instead of $\#\mathcal{E}$ actions, which is a massive reduction in dimensionality. However, evaluating  the objective and constraints by summing over all the entities is still very costly. To further reduce the computational cost, we can use a smaller set $\mathcal{S}$ of entities sampled uniformly at random from all entities processed by the optimization system in the previous period (e.g., the previous day) and use the mean reward models learned previously to estimate the expectations, leading to the optimization problem
\begin{align}
\label{eq:mpc_objective}
\min_w \sum_{e \in \mathcal{S}} \sum_{j=1}^{N_A} w_j \cdot \mu_{j}(x_e, a_e), & \quad \mbox{subject to} \\
\sum_{e \in \mathcal{S}} w_j \cdot \mu_{j}(x_e,a_e) \leq b_j, & \quad j = N_A+1, \ldots, N,
\end{align}
where $b_j = s  \cdot Budget_j$ is a rescaled budged where the scaling factor $s$ can be used to account for the relative size of the sample set $\mathcal{S}$ compared to the entire set $\mathcal{E}$, or to incorporate forecasted metric increases from one period to the next (e.g. due to a planned product change). To solve this optimization problem we use a grid search centered around the current metric tradeoff weights $w$ to generate candidate weight sets. We then set the new weights to be the candidate weight set that minimizes the abuse metrics while remaining within the budget constraints set on the cost metrics. 

%% file: section5_experiments.tex
\section{The PRO system in practice}\label{sec:system_in_practice}

The description of Predictive Response Optimization in Section~\ref{sec:algo} is generic; i.e., it can be applied to any \osn abuse problem using any set of abuse and cost metrics. In this section we turn our theory into reality, showing how to adapt the system to detect and block bots scraping an \osn.

We worked with product and engineering teams to implement the PRO system on \fbig\ifanon (which we will denote as \ig and \fb)\fi, both of which have more than one billion monthly active users. \ig is a ``directed'' social network where people follow creators and engage with their content, while \fb is an ``undirected'' social network where users connect and engage with people they know in real life.

On each \osn we implemented the system, collected data, trained a PRO model, and conducted online controlled experiments~\cite{abtest} to compare PRO's performance with that of a ``rule-based'' baseline. Due to differences in the ways users interact with the platforms as well as the state of each \osn's infrastructure, the exact baseline rules are specific to each \osn.
Below we describe our experimental setup, measure how much more abuse PRO can stop relative to our baseline system, and document observations about how PRO adapts to changing business, system, and adversarial conditions.

\subsection{Implementation Details}

\myparagraph{Metrics.} In order to implement PRO we first need to define the ``Abuse'' and ``Cost'' metrics introduced in Section~\ref{sec:opt}. We selected the following metrics for our experiment:

\begin{itemize}
\squeezelist
    \item \scraping: Count of logged HTTP requests identified as scraping using a scaled labeling system, with each request weighted by the number of units of user-identifiable information returned to the user. The labeling system consists of a set of rules generated by security analysts and expanded by automation. For example, one rule to detect a particular scraping attack is \texttt{user-agent="python-requests/<version>" and endpoint="<endpoint\_name>"}. We use this metric to quantify abuse.
    \item \daysactive: Count of calendar days during the measurement period during which an account is observed to be active. This metric correlates with user engagement and is used to quantify cost: if PRO is over-enforcing then \emph{days active} will decrease. (Due to our normalization and sign conventions~\eqref{eq:normalization}, the actual cost metric in our model will be ``loss of days active'' relative to the no-action baseline; i.e., $days\ active(a^0) - days\ active(a)$.)
    \item \feedback: Count of calendar days during the measurement period during which the account files a bug report. This metric correlates with incorrect actions and is used to quantify cost: if PRO takes too many actions on benign users, some users will perceive the enforcement to be a bug and \feedback will increase.
\end{itemize}

\noindent Note that each of these metrics can be calculated on a per-user basis to provide training data for the RL models. They can also be aggregated across multiple users for use in the MPC controller.

\myparagraph{Actions.}
At the time of our experiments, the following actions were available on both \osns:
\begin{itemize}
\squeezelist
    \item Temporarily disable the account. %
    \item Send the account through a compromise recovery flow.
    \item Invalidate all sessions, forcing the account to re-authenticate.
    \item Invalidate all sessions, plus limit the account to a single active session (i.e., each new login invalidates the existing session).
    \item Invalidate only the suspected automated session.
    \item No action.
\end{itemize}

\noindent In addition, the following actions were also available on \ig:
\begin{itemize}
\squeezelist
    \item Show a warning dialog that the user has to acknowledge before they can make further requests.
    \item ``Challenge'' the account by sending a One-Time Passcode (OTP) via SMS to the account's phone number.
    \item Show a CAPTCHA.
\end{itemize}

Each of these actions (other than ``disable'' and ``no action'') forces the user to perform some form of authentication to regain use of their account. Our hypothesis is that different actions will have different levels of effectiveness on abusive accounts (some may go away while others may pass the challenges and continue scraping) and different impacts for benign users (some  may pass the challenges and continue as before, while others may get frustrated and stop using the OSN entirely). The PRO system's goal is to optimize response selection based on the features of the account in question.

\myparagraph{Model training.}
When the system starts running, the RL models have no data. However, because we have a baseline rule-based system, we can initialize the RL model training dataset using features and metric values after the rule-based system takes action on the entities. Once the RL system starts taking actions, it logs training data based on its own actions and can be retrained daily.

We started training daily models 2--4 weeks before the experiments so that the RL system was in a steady state by the time the experimental results were collected. On \ig, our training data sets consisted of 8 million rows (accounts) and 201 columns (features). On \fb, our training data sets consisted of 8 million rows and 15 columns. On average, model training took 3.4 hours per metric on a 26-core x86 CPU with 64 GB RAM.

To assess accuracy of the model predictions, we measure the RL models' mean squared error (MSE) against ground-truth data. To normalize the MSE (i.e., squared $\ell_2$-norm of residuals) we divide it by the squared $\ell_2$-norm of the ground-truth values.  On \ig the normalized errors for \scraping, \daysactive, and \feedback are 0.51, 0.24, and 1.13, respectively, while on \fb the respective normalized errors are 0.51, 0.33, and 0.99.  These results show that our models for \scraping and \daysactive are good predictors. The relatively high error on \feedback is due to the high sparsity of the data: we see in Tables~\ref{table:results2} and \ref{table:results1} that fewer than three users out of every thousand file a bug report.

\subsection{Experimental setup}

\myparagraph{Experiment population.}
Both \ig and \fb were running a number of automation-detection classifiers $C_1, \ldots, C_K$ prior to the deployment of PRO, as well as classifiers for producing a general account-level ``abuse score''\ifanon\else~\cite{xu2021deep}\fi, which is used in the rule-based decision logic for \fb. Each classifier outputs a real-number score $s_i \in [0,1]$, and for each classifier we computed the threshold $t_i$ giving the classifier 90\% precision according to human-labeled ground-truth data.
On each \osn we then took a random sample of accounts for which {\em any} classifier score $s_i$ was greater than $t_i$ and assigned each of these accounts with probability 0.5 to either a Control group or a Test group. Accounts in the Control group received an action determined by a rule-based system (described below), while accounts in the Test group received an action determined by PRO.

For \ig, the experiment ran from Sep 25 to Oct 8, 2023 (\igdays days), and the metrics from accounts assigned to each group were cumulatively aggregated over the entire experiment period and compared.
546,289 unique accounts were selected for the Control group and 545,949 unique accounts were selected for the Test group.

For \fb, the experiment ran from Aug 7 to Oct 3, 2023 (58 days), and the metrics were cumulatively aggregated over the final \fbdays days (Sep 20 to Oct 3, 2023) and compared.
495,083 unique accounts were selected for the Control group, and 494,724 unique accounts were selected for the Test group.

At the time of the experiments, both \osns used manually designed, rule-based action-selection algorithms. Rule-based algorithms are state-of-the-art, used by various \osns (Section~\ref{sec:bg_enforcements}), and a multi-class classification baseline is not feasible due to the inability to obtain ground truth for which actions are optimal (Section~\ref{sec:intro}). Algorithm~\ref{alg:igrules} describes a representative example of the rules on \ig for this abuse scenario, while Algorithm~\ref{alg:fbrules} does the same for \fb. The algorithms incorporate the outputs of the automation classifiers $C_1,\ldots,C_K$ described above; in particular, we assume that these classifiers all output scores in $[0,1]$, with scores closer to 1 indicating higher likelihood of automation. Algorithm~\ref{alg:fbrules} also assumes we have an ``account abuse'' classifier\ifanon\else{ }such as that described in~\cite{xu2021deep}\fi.

\begin{algorithm}[t]\caption{Response selection logic for \ig}\label{alg:igrules}
\begin{algorithmic}
    \If{$\max(\{\mbox{automation classifier scores}\}) \ge  s_1$}
        \State disable account
    \Else \ no action
    \EndIf
\end{algorithmic}
\end{algorithm}

We note that while the automation classifiers used in the experiment are retrained throughout the experiment time periods, these classifiers are shared between control and test groups and thus affect both the baseline and PRO equally, enabling us to isolate the difference in performance between rule-based and PRO action selection in the experiment.

\begin{algorithm}[t]\caption{Response selection logic for \fb}\label{alg:fbrules}
\begin{algorithmic}
    \If{$\max(\{\mbox{automation classifier scores}\}) \ge  s_1$}
        \If{
          account abuse score $\ge s_2$ {\bf or} \\
          \quad \quad last compromise recovery was $\le N_1$ days ago {\bf or} \\
          \quad \quad account was registered $\le N_2$ days ago}
            \State disable account
        \Else
            \  send account through compromise recovery flow
        \EndIf
    \Else \ no action
    \EndIf
\end{algorithmic}
\end{algorithm}

\subsection{Experimental results}
\label{ss:results}

Experimental results for \ig and \fb appear in Table~\ref{table:results2} and Table~\ref{table:results1}, respectively. Metric values are summed across the last \fbdays days of each experiment and averaged per account.
We determined statistical significance using a two-sample $t$-test. Bold $p$-values indicate statistically significant results ($p \le 0.05$).

The experimental results show that our method can significantly reduce overall $Abuse$ metrics (see Equation~\eqref{eq:objective}). In particular, on \ig, {\bf PRO reduced \scraping by \igreduction} while causing no degradation in the two $Cost$ metrics, while
on \fb, {\bf PRO reduced \scraping by \fbreduction} with no statistically significant degradation in $Cost$ metrics.

The difference of an order of magnitude between the two experimental outcomes is a result of the system's development timeline. When we first began developing PRO on \ig, the heuristic rules on that \osn were fairly rudimentary, as evidenced by Algorithm~\ref{alg:igrules}. Rather than improve the rules, we focused our efforts on implementing and optimizing PRO and were able to realize the observed massive reduction in scraping relative to the prior state. During this period, the rules on \fb increased in sophistication, and when we turned our attention to the second \osn the rules were in the state exemplified by Algorithm~\ref{alg:fbrules}. Furthermore, the PRO system implemented for \fb is a simple port of the system developed and optimized for \ig; we expect that we could realize further gains with commensurate optimization. Thus the large difference in impact between the two \osns results from (a) a more sophisticated baseline algorithm on \fb and (b) less effort invested in optimizing PRO for \fb.\footnote{The fact that error metrics for model predictions are comparable on the two \osns suggests that the difference does \emph{not} result from the \ig models using more features than the \fb models.}

\input{results_table}

\subsection{Incorporating new business considerations}
\label{ss:business}

Another key feature of PRO is the ease of re-optimizing the system to incorporate new business considerations (cf.~examples in Section \ref{sec:business_constraints}). Here we share two case studies of introducing such considerations to PRO on \ig.

\myparagraph{Case Study 1: Reducing over-enforcement.}
After observing signs of over-enforcement\footnote{Since PRO is an optimization model rather than a classification model, the concepts of ``false positive'' and ``false negative'' do not technically apply to it. However, PRO can make locally sub-optimal decisions (as determined by information obtained later). We call actions that are sub-optimal in the cost dimension ``over-enforcement'' (corresponding to false positives) and actions that are sub-optimal in the abuse dimension ``under-enforcement'' (corresponding to false negatives).}
on the \ig \surface product, we formulated a new constraint aimed at limiting the reduction in user activity on \surface, in order to serve as a ``guardrail'' against over-enforcement. To implement this constraint we added to the PRO system a new cost metric quantifying {\em days active on \surface}.

\myparagraph{Case Study 2: Reducing SMS expenditure.}
Short message service (SMS) code verification is one of our available actions on \ig. Its goal is to verify user identity and/or prevent unauthorized access and abusive traffic coming from hacked accounts. Sending these codes has an associated financial cost. In order to reduce SMS expenditures, we added a new cost metric to PRO measuring {\smscost}.

\medskip \noindent
In both of the above cases, we adjusted the PRO system according to the following steps, and in both cases compared the effects of the two reward functions:

\begin{enumerate}
\squeezelist
  \item Log account-level attribution of the new cost metric.
  \vspace{-6pt}
  \begin{enumerate}
  \squeezelist
    \item For \emph{Case Study 1}, log whether the account is active on \surface each day.
    \item For \emph{Case Study 2}, log the total expenditure due to SMS messages sent to the account each day.
  \end{enumerate}
  \vspace{-4pt}
  \item Join the cost-attribution logs with enforcement-action logs and account features to generate training data for new metric-prediction models (Section \ref{sec:reward_prediction}).
  \item Train the new metric-prediction models.
  \item Solve~\eqref{eq:mpc_objective} to determine the metric weights in the reward function of Equation~\eqref{eq:reward}, with the new constraint $Budget_k$ as one of the algorithm inputs.
  \item Update the system's reward function with the new metric weights and new metric prediction models.
\end{enumerate}

In \emph{Case Study 1}, we determined that the product impact of over-enforcement was significant enough to warrant an immediate model adjustment rather than an online controlled experiment; we therefore used a ``before and after'' approach to quantify the impact. We collected data on the previous reward function for the 7 days from Jun 20 to 26, 2023, launched the new reward function on Jun 27, and collected data from Jun 29 to Jul 5. We found that the new reward function increased \emph{days active on \surface} by \textbf{\mwebdaulossreduction} ($p = 0.02$) and decreased \scraping by \textbf{\mwebscrapingreduction} ($p = 0.006$), showing that re-weighting the reward function can both reduce cost and increase effectiveness.

In \emph{Case Study 2}, we ran an online controlled experiment, using the new reward function in the Test group and the previous reward function in the Control group.
We collected data from Sep 17 to 23, 2023 (7 days) comparing 1,277,330 accounts in Control with 1,277,823 accounts in Test.\footnote{Since this experiment involved changing PRO's metric weights, which have a smaller effect than comparing PRO with a baseline selection algorithm, we increased the size of the experiment in order to ensure statistical significance.}
The data show that we reduced \smscost by \textbf{\smsreduction} ($p \ll 0.001$) without any significant impact on \scraping ($2.0 \pm 2.6\%$ reduction; $p=0.13$). Qualitatively, we observed that PRO switched to other available enforcement actions of similar effectiveness whenever possible, reserving SMS code verification for entities where it would be most effective at stopping abuse.

\subsection{Onboarding new enforcement actions}
\label{ss:new_action}
The PRO system simplifies the process of introducing and testing new enforcement actions. In the absence of an ML-based system to select enforcement actions, action selection relies on domain expertise to create hard-coded rules that decide when to apply the new enforcement action. RL, on the other hand, addresses this ``cold-start problem'' via exploration.

\begin{figure*} %
\centering
\begin{subfigure}[t]{0.33\textwidth}
\includegraphics[width=\textwidth]{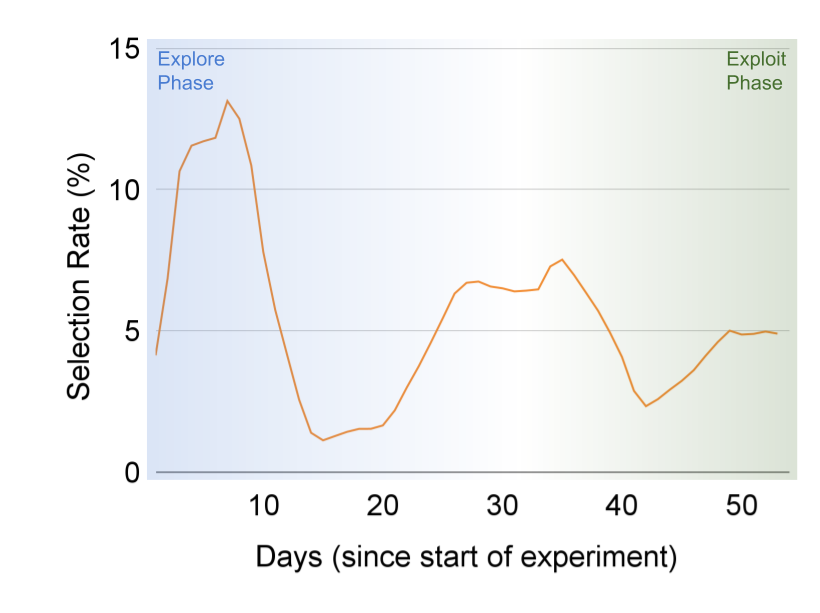}
\caption{Selection Rate}\label{fig:chart_4_2}
\end{subfigure}
\begin{subfigure}[t]{0.33\textwidth}
\includegraphics[width=\textwidth]{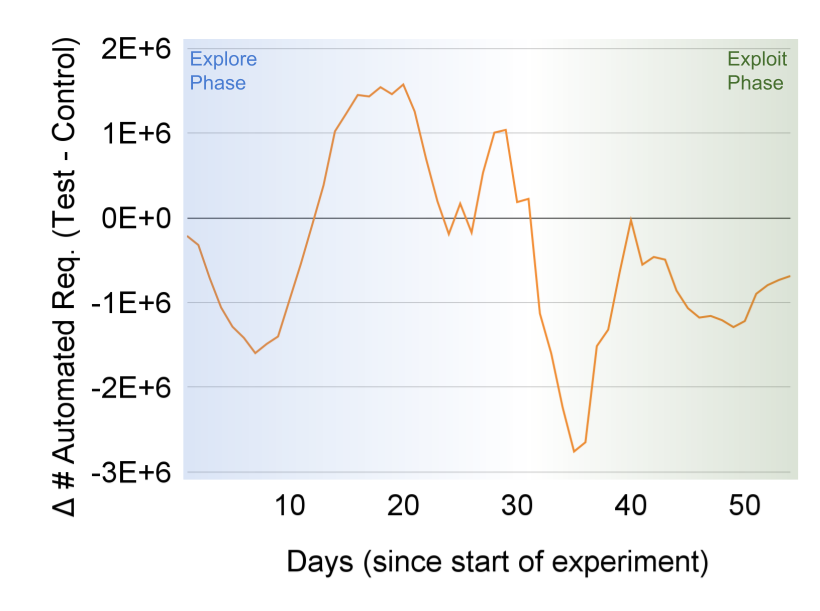}
\caption{$\Delta$ Automated Request Count}\label{fig:chart_4_1}
\end{subfigure}
\begin{subfigure}[t]{0.33\textwidth}
\includegraphics[width=\textwidth]{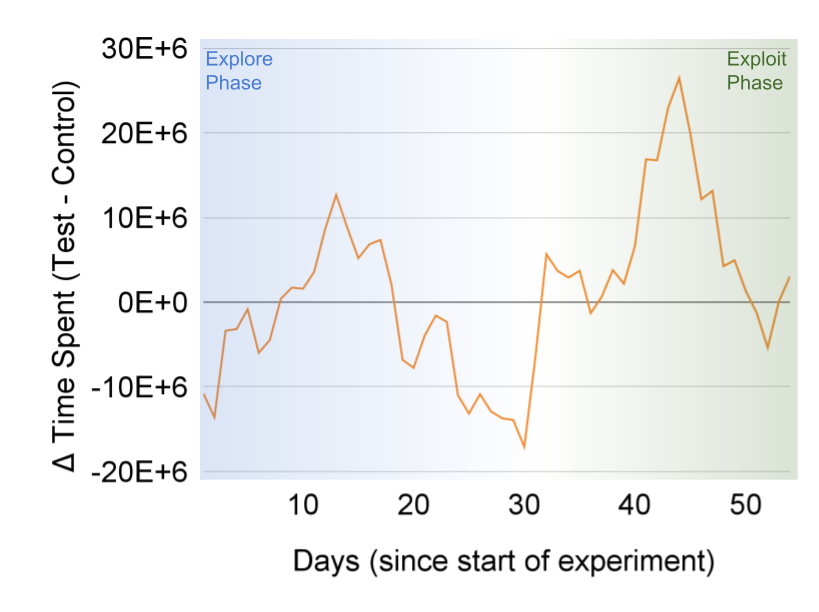}
\caption{$\Delta$ Time Spent}\label{fig:chart_4_3}
\end{subfigure}
\captionsetup{justification=centering,margin=1cm}
\caption{(a) Selection rate of the new enforcement action. (b) Daily deltas ($Test-Control$) of the abuse metric \automation. (c) Cost metric {\em time spent} (7-day moving average)}\label{fig:adding_action}
\end{figure*}

\myparagraph{Case Study 3: A new enforcement action.} We implemented a new enforcement action on \fb and added it to the ``library'' of PRO actions. The action invalidates existing web sessions created by the account, forcing the account to re-authenticate. In addition, it restricts the account from creating multiple concurrent sessions, allowing only one device to be logged in to \fb{} at any given time. Our hypothesis was that the new action would be more effective against accounts that use multiple concurrent sessions to perform automation, while having smaller impact on incorrectly classified users than the account-disable action, since one session is still allowed. We ran experiments with this action using the following metrics:

\begin{itemize}
\squeezelist
    \item \automation (abuse metric): Number of HTTP requests identified to be resulting from scraping (i.e., unweighted version of \scraping).
    \item \timespent (cost metric): Time duration (in seconds) that the account spends active on the \osn (i.e., continuous version of \emph{days active}).
\end{itemize}

In Figure~\ref{fig:adding_action}, we see that initially PRO does not have any knowledge about the potential impact of the new enforcement action. In this stage we observe large fluctuations in selection rate (\ref{fig:chart_4_2}), accompanied by overall sub-optimal action selection with higher numbers of \automation~(\ref{fig:chart_4_1}) and lower user-engagement metrics (\ref{fig:chart_4_3}). Around day 31, the system stabilizes with smaller shifts in action-selection rate, more optimal action selection (i.e., reduction in abuse) in the test group, and higher user-engagement metrics. At this point, the system is starting to utilize the new enforcement action more effectively.

At the end of the experiment, we aggregated metrics from the final 5 days (Aug 25 to 29, 2023) comparing 1,057,156 accounts in Control with 1,055,797 accounts in Test. Results showed the new action led to a {\bf \newresponsereduction} ($p=0.008$) reduction in \automation, and no statistically significant change in \timespent ($0.7 \pm 1.6\%$ reduction; $p=0.38$).

\subsection{Uncontrolled systemic changes}
\label{ss:adaptation}

Since our main results in Section~\ref{ss:results} are based on data aggregated over a two-week period, an important question is how the system reacts to changes in feature distributions over longer periods of time (``concept drift''). We expect the PRO system to adapt automatically to such changes since we are retraining the models daily; here we share two case studies supporting this claim.

\myparagraph{Case Study 4: Automatically adjusting to a bug.} Engineers inadvertently introduced a bug into an identity-verification challenge on \ig that asked account owners to upload photos of their face. This challenge was previously found to be effective at stopping abuse stemming from automated activity. However, the bug caused some enrolled accounts to remain in a ``stuck'' state with no ability to clear the challenge. After the bug manifested, new observed data points showed that the action had a significant negative impact on $Cost$ metrics, which led model coefficients to change significantly after the model was retrained. As a result, PRO completely stopped selecting this action two days after the bug was observed, without engineers manually altering the configuration to disallow the action from being selected.

\myparagraph{Case Study 5: Adjusting to adversarial adaptation.} We incorporated a new ``warning'' challenge into the PRO system's action suite on \ig. This challenge presents accounts suspected of automated activity with a warning notice and prevents any future web requests from being served until the account acknowledges the warning. We ran an online controlled experiment and aggregated metrics from Mar 24 to Apr 6, 2023 (14 days), comparing 803,813 accounts in Control with 803,753 accounts in Test. Results showed the addition of this new challenge led to a \textbf{15\%} reduction ($p \ll 0.001$) in \scraping and no statistically significant change in \timespent ($0.1 \pm 0.3\%$ increase; $p=0.50$). In the Test group, PRO selected the new action \textbf{13\%} of the time.

Based on these promising results, we increased the size of the experiment group. A month later (Apr 25 to May 8, 2023) we observed that the daily selection rate for the new action had a statistically significant drop ($p \ll 0.001$), falling to only \textbf{4\%}. Data analysis revealed that traffic from some abusive entities resumed almost immediately after the warning challenge was presented to them, providing evidence that some adversaries had learned how to circumvent the challenge and the system needed to select other, more effective responses to stop them. Despite this adversarial adaptation, we found that overall the 14-day rolling average of \scraping decreased by 24\% ($p = 0.006$) between Apr 6 and May 8, indicating that our changes did have beneficial impact on the overall scraping ecosystem.

%% file: results_table.tex
\begin{table}%
\resizebox{\columnwidth}{!}{%
\begin{tabular}{ccccc}
\toprule
Metric & Control & Test & Delta & $p$-value \\ 
\midrule
\emph{weighted scraping}
    & $16,700$ & $\mathbf{6,830}$ & $\mathbf{-59.2\%}$ & $\mathbf{0.00}$ \\ 
\emph{requests} \\
\daysactive & $2.94$ & $\mathbf{2.98}$ & $\mathbf{+1.4\%}$ & $\mathbf{6.3 \times 10^{-36}}$ \\
\feedback & $2.50 \times 10^{-3}$ & $2.80 \times 10^{-3}$ & $+12.0\%$ & $0.089$ \\
\bottomrule
\end{tabular}
}
\caption{Experimental results for \ig{}. Reported numbers are per-account averages over a \igdays-day period.}
\label{table:results2}
\end{table}

\begin{table}%
\resizebox{\columnwidth}{!}{%
\begin{tabular}{ccccc}
\toprule
Metric & Control & Test & Delta & $p$-value \\ 
\midrule
\emph{weighted scraping}
    & $3,540$ & $\mathbf{3,380}$ & $\mathbf{-4.51\%}$ & $\mathbf{0.0395}$ \\ 
 \emph{requests} \\
\daysactive & $3.26$ & $3.25$ & $-0.430\%$ & $0.0777$ \\
\feedback & $1.20 \times 10^{-3}$ & $9.61 \times 10^{-4}$ & $-20.2\%$ & $0.163$ \\
\bottomrule
\end{tabular}
}
\caption{Experimental results for \fb{}. Reported numbers are per-account averages over a \fbdays-day period.}
\label{table:results1}
\end{table}

%% file: section7_ethics.tex
\section{Ethics considerations}
\label{sec:ethics}

We have assessed the value of publishing this work against potentially adverse consequences due to its methodology and/or data practices. Specifically, we considered risks related to \emph{user harm}, \emph{equitable selection}, \emph{user consent}, and \emph{user data handling}~\cite{irb}.

\myparagraph{User harm.}
We acknowledge that fighting abuse on online social networks is a task fraught with risk: some users are harmed by the abuse itself, while others are harmed by over-enforcement when defenses become too aggressive. This entire work is devoted to systematically balancing reduction in both of these harms. Our system's ``explore-exploit'' strategy aims to reduce harm over the entire platform while taking into account that not all local decisions can be perfect. However, our experiments demonstrate that the PRO system offers significant benefit to users in terms of overall harm reduction.

We also considered the risk of our experiments over-enforcing on benign users due to system and/or model deficiencies. To mitigate this risk we set up our experiments to include only users that our binary classifiers predicted to be abusive with high confidence. The precision of the binary classifiers was confirmed to be greater than 90\% at the time of the experiments. Given the above considerations, we believe that our experiment exposes users to risks that are ``reasonable in relation to anticipated benefits''~\cite{irb}. 

\myparagraph{Equitable selection.} Our subjects are necessarily chosen equitably since a user's participation in the experiment is determined by a random number generator whose output is independent of any user properties.

\myparagraph{User consent.}
Partridge and Allman~\cite{partridge-allman} observe that ``direct
consent is not possible in most Internet measurements,'' and our study is a good example.
Since we cannot predict in advance who our classifiers will determine to be abusive, we would have to obtain consent either from the entire \osn population or from the specific users acted on by PRO at the moment action is taken. Building a bespoke consent flow for this study would be a large engineering task and would risk both information bias (users might act differently knowing they were in a security study) and selection bias (both benign and abusive users who choose to participate in a security study may not reflect the general population). Either type of bias could render our statistical analyses invalid and thus handicap our ability to measurably improve abuse detection.

Partridge and Allman suggest that ``proxy consent'' is the \emph{de facto} standard in large-scale internet measurement studies, giving the example of ``network measurements taken on a university campus typically seek consent from the university.'' In our case institutional consent was rendered via the \osns' agreements to let us conduct and publish this research. In particular, the reviewers approving the research noted that all users in our study have accepted the terms of service of \ig or \fb (as applicable), which address use of data in the context of investigating suspicious activity and addressing policy violations.
The reviewers therefore concluded that the \osn terms of service provide users a sufficient level of transparency.

We note that explicit consideration of user consent is not historically an element of large-scale internet security studies. Bilogrevic \emph{et al.}~\cite{bilogrevic-google} use a proxy approach similar to ours, deriving their user consent from the fact that users opted in to a setting to ``Make searches and browsing better.'' However, recent work studying millions of users on Reddit~\cite{kumar-reddit}, Facebook~\cite{golla-facebook,onaolapo-facebook}, and Google Chrome~\cite{thomas-stuffing} do not address user consent at all in their ethics discussions. We discuss open questions in this area in Section~\ref{sec:conclusion}.

\myparagraph{Data handling.}
Before developing and testing our system,
we assessed how data would be used and protected and ensured that technical systems and/or manual processes were in place to mitigate any identified risks prior to the launching the experiment\ifanon\else~\cite{meta-privacy-review}\fi. For this project, we mitigated risks by:
\begin{itemize}
    \squeezelist
    \item Limiting data collection to a set of user features identified as being relevant for abuse detection;
    \item Specifically excluding any sensitive data from collection;
    \item Restricting access to both collected and inferred data;
    \item Deleting all user-identifying data within 90 days of collection;
    \item Using technical safeguards to ensure that the data are only used for safety, integrity, and security use cases.
\end{itemize}

%% file: section6_conclusion.tex
\section{Directions for Future Work}
\label{sec:conclusion}

\myparagraph{Optimizing long-term reward}. PRO selects actions to maximize reward over a fixed time horizon. However, we may wish to select actions to reduce abuse and cost in the long term. We could view the RL problem as a ``continuing'' task, or as an infinite-horizon task instead of a finite-horizon one~\cite{sutton2018reinforcement}, and optimize using RL algorithms such as Q-learning \cite{watkins1992q,mnih2015human,kalweit2020deep}. For each metric, we could train on the sequence of actions taken on each entity and leverage Q-learning's ability to learn cumulative long-term discounted rewards.

\myparagraph{Evaluating exploration strategies}. Unlike supervised learning, RL involves an explore-exploit tradeoff. If PRO always issues the same action to an entity, it can never learn whether that action was actually the best choice. Having this feedback loop is even more important in an adversarial setting. However, comparing exploration strategies can be challenging. Simply A/B testing the same RL system with two exploration implementations will not work if the models share training data, as the Test model will ``free-ride'' on the Control model's exploration, or vice versa. Developing approaches to compare exploration methods would allow us to test other exploration strategies such as Upper Confidence Bound~\cite{carpentier2011upper} or Quantile Regression-based sampling~\cite{quantileregression}.

\myparagraph{Addressing over- and under-enforcement.} Case Study 1 describes one set of users on which the system made locally sub-optimal decisions. Other such populations include:
\begin{itemize}
    \squeezelist
    \item ``Power users,'' defined as non-malicious users who use the platform in such a way that their activity appears automated. The population of power users is so small that over-enforcement on this subset doesn't meaningfully impact the global cost metrics.
    \item ``Repeat offenders,'' defined as users sent through the PRO system multiple times. If the system doesn't update features quickly enough then it risks repeating a response that was either too harsh for a benign user or not effective on a malicious user.
    \item ``Low-information'' users, who may use unauthorized third-party tools or otherwise inadvertently breach the \osn terms of service.
\end{itemize}
For each of these cases, we believe that some combination of new cost metrics (as in Case Study 1) and/or new enforcement actions (as in Case Study 3) can improve the model's performance.

\myparagraph{Generalizing our solution to other anti-abuse use cases.} Our experiments provide empirical evidence about our system yielding measurable improvements in reducing scraping of \osns. Assessing how our solution can impact other abuse problems remains an open area of research.

\myparagraph{Understanding consent in adversarial studies.} In our discussion of user consent we asserted that ``users might act differently knowing they were in a security study'' and that ``both benign and abusive users who choose to participate in a security study may not reflect the general population.'' These assertions have never been tested rigorously; a study that tested hypotheses on user consent in adversarial environments would provide crucial scientific input to the ethical standards for all future studies  involving real-world adversaries.

\ifanon
\else

\newpage

\subsection*{Authors' Contributions}

\begin{itemize}
    \squeezelist
    \item Garrett Wilson developed model enhancements and observability infrastructure, drafted the technical portions of this paper, and coordinated the paper-revision process.
    \item Geoffrey Goh designed the experimentation process and developed core components including metrics, enforcement actions, serving infrastructure, and the model predictive controller.
    \item Yan Jiang extended PRO to \fb and ran related experiments for this paper.
    \item Ajay Gupta and Jiaxuan Wang provided engineering support.
    \item David Freeman consulted in the initial design phase of PRO and coordinated the writing process.
    \item Francesco Dinuzzo created this project and led the team that built and maintained the PRO system. He designed the reinforcement learning approach to abuse mitigation, built the initial production versions of the system, and guided its development over multiple iterations.
\end{itemize}

\subsection*{Acknowledgments}

The authors thank Katriel Cohn-Gordon, Feargus Pendlebury, Francesco Logozzo, Chris Palow, and Yiannis Papagiannis for helpful feedback on earlier drafts of this paper. We thank Sandeep Hebbani, Emile Litvak, and Gemma Silvers for encouraging publication of this paper. We thank the five anonymous reviewers for their feedback, and in particular the shepherd who helped us craft the paper into its current form.

\fi